\documentclass[conference]{IEEEtran}
\IEEEoverridecommandlockouts

\usepackage[switch]{lineno}  

\usepackage{cite}
\usepackage{amsmath,amssymb,amsfonts}
\usepackage{graphicx}
\usepackage{textcomp}
\usepackage{xcolor}
\def\BibTeX{{\rm B\kern-.05em{\sc i\kern-.025em b}\kern-.08em
    T\kern-.1667em\lower.7ex\hbox{E}\kern-.125emX}}

\usepackage[hidelinks]{hyperref}  

\usepackage{amsmath,amssymb,amsfonts}
\usepackage{graphicx}
\usepackage{textcomp}
\usepackage[dvipsnames]{xcolor}

\usepackage{tabularx,booktabs,array}
\usepackage{arydshln}

\usepackage{algorithm}      
\usepackage{algpseudocode}  

\DeclareMathOperator{\tr}{trace}

\begin{document}

\title{SATORIS-N: Spectral Analysis based Traffic Observation Recovery via Informed Subspaces and Nuclear-norm minimization}


\author{\IEEEauthorblockN{Sampad Mohanty}
\IEEEauthorblockA{\textit{Thomas Lord Department of Computer Science} \\
\textit{University of Southern California}\\
Los Angeles, USA \\
sbmohant@usc.edu}
\and
\IEEEauthorblockN{Bhaskar Krishnamachari}
\IEEEauthorblockA{\textit{Ming Hsieh
Department of Electrical and Computer Engineering} \\
\textit{University of Southern California}\\
Los Angeles, USA \\
bkrishna@usc.edu}
}
\maketitle

\begin{abstract}
Traffic-density matrices from different days exhibit both low rank and stable correlations in their singular-vector subspaces. Leveraging this, we introduce \textbf{SATORIS-N}, a framework for imputing partially observed traffic-density by informed subspace priors from neighboring days. Our contribution is a \textbf{subspace-aware semidefinite programming (SDP)} formulation of nuclear norm that explicitly informs the reconstruction with prior singular-subspace information. This convex formulation jointly enforces low rank and subspace alignment, providing a single global optimum and substantially improving accuracy under medium and high occlusion. We also study a lightweight \textbf{implicit subspace-alignment} strategy in which matrices from consecutive days are concatenated to encourage alignment of spatial or temporal singular directions. Although this heuristic offers modest gains when missing rates are low, the explicit SDP approach is markedly more robust when large fractions of entries are missing. Across two real-world datasets (Beijing and Shanghai), \textbf{SATORIS-N} consistently outperforms standard matrix-completion methods such as SoftImpute, IterativeSVD, statistical, and even deep learning baselines at high occlusion levels. The framework generalizes to other spatiotemporal settings in which singular subspaces evolve slowly over time. In the context of intelligent vehicles and vehicle-to-everything (V2X) systems, accurate traffic-density
 reconstruction enables critical applications including cooperative 
perception, predictive routing, and vehicle-to-infrastructure (V2I) communication optimization. When 
infrastructure sensors or vehicle-reported observations are incomplete—
due to communication dropouts, sensor occlusions, or sparse connected
vehicle penetration—reliable imputation becomes essential for safe and 
efficient autonomous navigation.
\end{abstract}


\section{Introduction}

Modern intelligent transportation systems increasingly rely on vehicle-to-everything (V2X) communication networks to enable cooperative perception, predictive routing, and autonomous navigation. In these systems, accurate traffic-density reconstruction is essential for safe decision-making, yet observations from connected vehicles and vehicle-to-infrastructure (V2I) sensors are frequently incomplete. Communication dropouts, sensor occlusions, and sparse connected-vehicle penetration rates (typically 10-30\%) create substantial missing-data challenges. Even 10\%--35\% missing entries can eliminate 35\%--98\% actionable information~\cite{Sun2022-kq}, severely impairing downstream analytics and control systems. Reliable imputation methods are therefore critical for operational V2X networks. Classical matrix-completion methods exploit low-rank structure but ignore temporal consistency across observation periods. Previous work~\cite{ANRGPCA} and our spectral analysis demonstrate that daily traffic-density matrices constructed from spatial grids and hourly intervals exhibit not only strong low-rank structure but also highly stable singular-vector subspaces across consecutive days. These properties suggest that cross-day subspaces provide valuable priors for imputing missing traffic data. Theory indicates that known subspaces can reduce sample complexity~\cite{PriorMaxCorr2018}, motivating subspace-informed recovery for traffic and other spatiotemporal datasets where temporal correlations persist. For intelligent vehicle systems with sparse connected-vehicle observations or intermittent V2I sensor data, this subspace-informed approach enables accurate traffic reconstruction for cooperative perception and predictive routing applications.

Motivated by this, we propose \textbf{SATORIS-N}, a subspace informed low rank imputation framework. We devise an implicit subspace alignment meta-algorithm which applies any standard matrix imputation algorithm on concatenated adjacent-day matrices to encourage shared spatial/temporal subspace which are then used as baselines. Our primary contribution is an explicit \textbf{subspace-aware semidefinite program (SDP)} that injects dominant singular vectors from reliable neighboring days into a convex nuclear-norm formulation. This yields reconstructions that fit observed entries, enforce low rank, and align with informed subspaces, guaranteeing a global optimum and avoiding non-convex sensitivity of tensor models. Experiments on Beijing and Shanghai taxi-GPS datasets show that SATORIS-N consistently outperforms mean/median fills, $k$-NN, SoftImpute, MICE, and deep-learning based methods across 10\%--90\% missingness. Under the high occlusion regime (75\% -- 90\% missing), explicit SDP reduces relative root mean square error (RRMSE) and mean absolute error (MAE) by 10\% -- 20\% relative to the best baseline, while deep models significantly degrade in performance due to their higher sample complexity.
The main contributions of this paper are as follows:

    \textbf{(1) Implicit and Explicit Subspace-Aware Formulations.}
    We propose both an implicit alignment meta-algorithm and an explicit subspace-aware SDP. The implicit variant promotes shared subspaces via adjacent-day stacking and promotes any base subspace-unaware imputation algorithm into a subspace-aware one. The explicit convex nuclear-norm SDP injects known subspaces as priors, admits a single global optimum due to convexity, and achieves the best performance in comparison to ten reference implicitly aware baselines when occlusion levels are severe (75\% - 90\%)

     \textbf{(2) Quantification of Cross-Day Subspace Stability.}
    Extending our prior spectral analysis, we quantify subspace stability in multi-week Shanghai data, showing that rank-$k$ ($k \le 10$) common subspaces capture over 75\% of the variance across three consecutive days, validating neighboring-day subspaces as reliable priors.

     \textbf{(3) Extensive Empirical Evaluation.}
    We compare implicit and explicit formulations against statistical, low-rank, and deep-learning baselines over 10--90\% missingness. Explicit SDP variants consistently achieve lower errors under medium-to-high occlusion and remain competitive even at extreme sparsity.

     \textbf{(4) Reproducible Code and Unified Interface.}
    We release open-source implementations with a lightweight command-line interface and \texttt{scikit-learn}-style \texttt{fit}/\texttt{transform} APIs~\cite{sklearn_api} supporting both implicit and explicit modes, enabling adoption in spatiotemporal analytics pipelines. Especially, the implicit mode accepts any base algorithm supporting sklearn-api and makes it subspace-aware.



\section{Related Prior Work}
\label{sec:related}

Traffic-data imputation has been extensively studied using low-rank matrix and tensor models. Early work by Asif et al.~\cite{Asif2016Matrix} applied matrix factorization and tensor decomposition to large-scale road-network data, showing that exploiting low-dimensional structure can significantly improve the accuracy of imputation. This low-rank assumption is supported by empirical observations that traffic matrices exhibit strong temporal and spatial dependencies due to recurring daily patterns~\cite{Qu2009PPCA, Li2013Efficient}. Building on this foundation, Chen et al.~\cite{Chen2020NonconvexTensor} proposed a nonconvex low-rank tensor completion method for spatiotemporal traffic data, which captures underlying patterns more effectively and achieves higher accuracy than conventional convex approaches. Other traffic-imputation methods further exploit low-rank structure or auxiliary signals, including ensemble correlation-based matrix completion~\cite{chen2017eclrmc}, improved low-rank matrix decomposition that incorporates missing-value patterns~\cite{luo2019ilrmd}, truncated nuclear-norm tensor models~\cite{Chen2020NonconvexTensor}, tensor-ring decomposition with nonlocal subspace regularization~\cite{wu2023trnlsr}, robust Tucker factorization~\cite{lyu2024tucker}, knowledge-graph–driven generative priors~\cite{liu2024kggan}, and latent probabilistic models such as PPCA~\cite{huang2023ppca}. Some of these approaches implicitly leverage correlated subspaces by stacking or tensorizing multi-day traffic data, thereby inducing shared latent structures across days. However, they provide little control over the strength or direction of subspace alignment and do not explicitly encode historical singular subspaces as priors. In contrast, our framework is convex by design and directly incorporates subspace information within the optimization. As shown by Ding and Udell~\cite{ding2021simplicity}, many low-rank matrix recovery problems admit robust and well-conditioned convex SDP formulations, and related approaches based on SDP plus low-rank have been applied in diverse domains—including synchronization, community detection, phase retrieval, and robust localization~\cite{pmlr-v49-bandeira16,huang2017phaselift,foucart2019binary,clark2023smile}.

Deep generative models have also been explored to handle high missing rates and incorporate external context. For example, KG-GAN~\cite{liu2024kggan} uses a knowledge graph–enhanced GAN that integrates points of interest, weather, and other contextual signals into the imputation process, improving robustness under severe missingness. While such deep models can capture complex nonlinear spatiotemporal patterns, they typically require large amounts of historical data and may overfit highly variable traffic flows~\cite{yu2018spatio}. They are also data-hungry and computationally intensive, incurring high training costs for large-scale networks~\cite{zheng2020gman}, and often act as black boxes with limited interpretability compared to matrix-based or statistical methods.

In contrast, \emph{SATORIS-N} formalizes both \emph{implicit} and \emph{explicit} subspace mechanisms within a unified convex, matrix-based framework. The implicit variant aligns adjacent-day matrices via simple concatenation along shared spatial or temporal dimensions to encourage common modes, while the explicit variant embeds estimated singular-subspace priors directly into a semidefinite program. This direction aligns with theoretical work on correlation-based priors for structured recovery~\cite{PriorMaxCorr2018}, which we extend to spatiotemporal traffic data with practical subspace estimation and imputation routines. The resulting SDP based convex formulation ensures a single global optimum, avoids the nonconvex pitfalls of tensor models, and remains effective under sparse observations where deep networks tend to degrade. 
 To our knowledge, no prior traffic-imputation method explicitly encodes historical subspace structure within an SDP based convex optimization formulation, distinguishing \emph{SATORIS-N} from traditional low-rank approaches and data-hungry deep-learning models.

\section{Problem Formulation and Methodology}
\label{sec:problemformulation}

\begin{table}[h]
\caption{Summary of Notation}
\centering
\small
\begin{tabular}{p{0.30\linewidth} p{0.60\linewidth}}
\hline
\textbf{Symbol} & \textbf{Description} \\
\hline
$m,n,p,k$ &
Spatial grid dimensions $m\times n$, number of time slots $p$ per day, and target rank $k\le \min\{mn,p\}$. \\[2pt]

$D[t]\in \mathbb{R}^{mn\times p}$ &
Traffic-density matrix for day $t$, formed by vectorizing the $m\times n$ spatial grid into $mn$ rows; full observations, nothing missing. \\[2pt]

$X\circ Z$ &
Hadamard (element-wise) product of conformable matrices $X$ and $Z$. \\[2pt]

$\Omega[t]\in\{0,1\}^{mn\times p}$ &
Observation mask for day $t$. The projection $P_{\Omega[t]}(X)=X\circ \Omega[t]$ keeps observed entries and zeros the rest. \\[2pt]

$Y[t]=P_{\Omega[t]}(D[t])$ &
Partially observed data for day $t$. \\[2pt]

$\|X\|_{*},\ \|X\|_{F}$ &
Nuclear and Frobenius norms, respectively. \\[2pt]

$\operatorname{tr}(X)$ &
Trace of a square matrix $X$. \\[2pt]

$\mathbb{S}_+^d,\ X\succeq 0$ &
Cone of $d\times d$ symmetric positive semidefinite matrices; $X\succeq 0$ means $X\in \mathbb{S}_+^d$. \\[2pt]

$t_1,\ t_2$ &
Target-day index and neighboring/adjacent-day index. \\[2pt]

$U[t_2]\Sigma[t_2]V[t_2]^\top \approx D[t_2]$ &
Rank-$k$ truncated singular value decomposition (SVD) with $U[t_2]\in\mathbb{R}^{mn\times k}$, $\Sigma[t_2]\in\mathbb{R}^{k\times k}$ diagonal, $V[t_2]\in\mathbb{R}^{p\times k}$, and $U[t_2]^\top U[t_2]=V[t_2]^\top V[t_2]=I_k$. \\[2pt]

$u_i^{(t_2)},\ v_i^{(t_2)}$ &
$i$-th columns of $U[t_2]$ and $V[t_2]$ ($i=1,\dots,k$). \\[2pt]

$\hat D[t_1]\in\mathbb{R}^{mn\times p}$ &
Imputed (completed) matrix for day $t_1$. \\[2pt]

$S\in\mathbb{R}^{k\times k}$ &
Coefficient matrix with $\hat D[t_1]=U[t_2]\,S\,V[t_2]^\top$. Since $U[t_2],V[t_2]$ have orthonormal columns, $\|\hat D[t_1]\|_*=\|S\|_*$. \\[2pt]
\hline
\end{tabular}
\label{tab:notations}
\end{table}

\subsection{SATORIS-N: Assumptions}

Our approach relies on four empirically supported properties of daily traffic-density matrices. 
(1) \textbf{Low rank:} each $\mathbf{D}[t]$ admits a strong rank-$k$ approximation capturing $>75\%$ variance~\cite{ANRGPCA,Lakhina2005Structural,Soule2005Traffic,Asif2016Matrix,Asif2013LowDimRoad,Li2013Efficient,Qu2009PPCA}. 
(2) \textbf{Cross-day stability:} dominant spatial/temporal subspaces change little across adjacent days, i.e., principal angles between $\mathbf{U}[t_1],\mathbf{U}[t_2]$ and $\mathbf{V}[t_1],\mathbf{V}[t_2]$ remain small for $|t_1-t_2|\le\tau$. 
(3) \textbf{Neighbor-day availability:} at least one nearby day with near-complete data permits reliable truncated-SVD (Singular Value Decomposition) subspace estimates. 
(4) \textbf{MAR/MCAR:} Missing At Random (MAR) / Missing Completely At Random assumption : the observed entries provide an unbiased sample of $\mathbf{D}[t_1]$. 
Section~\ref{sec:experiments} validates these assumptions and shows that subspace-aware imputation yields substantial gains.

\paragraph*{Motivation for Implicit Priors}
Under the preceding assumptions, the dominant spatial and temporal
singular subspaces of daily traffic matrices remain strongly correlated
across neighboring days. This naturally motivates \emph{implicit subspace
priors}, where information from adjacent days is reused to guide completion
of a partially observed matrix. Before introducing our explicit convex
formulation, we therefore describe a simple heuristic baseline that
encourages such alignment via direct matrix stacking across days. Although
it does not impose formal subspace constraints, this approach provides a
reference point for evaluating explicit subspace-aware optimization and
acts as a meta-algorithm that can improve the performance of any standard
matrix-based imputation method.

\subsection{Implicit Subspace Alignment}
To heuristically exploit the coherence of the inter-day subspace without modifying
the optimization objective/algorithm, we concatenate a target matrix $D[t_1]$ with
an adjacent, mostly complete matrix $D[t_2]$ and apply a standard low-rank
completion algorithm to the stacked matrix.

We consider two stacking modes:
\begin{itemize}
    \item \textbf{Horizontal stacking (H):}
    $\tilde{D} = [\, D[t_1] \; D[t_2] \,]$ aligns the matrices along
    columns, promoting shared \emph{spatial} (left) singular subspaces.
    \item \textbf{Vertical stacking (V):}
    $\tilde{D} = [\, D[t_1]^\top \; D[t_2]^\top \,]^\top$ aligns them along
    rows, promoting shared \emph{temporal} (right) singular subspaces.
\end{itemize}

Any standard matrix-completion algorithm (e.g., Soft-Impute or IterativeSVD)
is then applied to $\tilde{D}$, and the recovered block corresponding to
$D[t_1]$ is extracted for evaluation. These two variants, denoted
\texttt{-h} and \texttt{-v}, serve as implicit subspace-alignment baselines
and are used throughout our experiments to quantify the benefits of implicit
versus explicit subspace injection in SATORIS-N. We refer to the implicitly
informed nuclear-norm variant as \emph{Soft Reconstruction Implicit subspace
injection} (\texttt{SRISI}), also called \texttt{NNmin-h}; the connection
to the explicit formulation will become clear in the next section.

\subsection{Explicit Subspace Alignment: Subspace Informed Nuclear Norm Minimization via SDP}

Although implicit alignment by stacking matrices gives us a starting point, it has a major drawback - stacking allows us to share only one subspace - either the spatial or the temporal, not both simultaneously. A weighted average of the values imputed by stacking horizontally and vertically could be taken and may slightly alleviate the problem in post processing. This leads us to imputation based on explicitly informed priors using semi-definite-programming.

The nuclear norm of a matrix is defined as the sum of its singular values which can be easily obtained via the singular value decomposition. $||X||_*$ is also given by $\tr(\sqrt{X^TX})$. The nuclear norm, being a proper mathematical norm, is a convex function. It turns out that there are various ways to formulate the nuclear norm as the solution to an optimization problem, for example, (\ref{eq:nnform1}) and (\ref{eq:nnform_sdp}).

\begin{equation}
\label{eq:nnform1}
    \min_{A,B} \frac{1}{2} (||A||_F^2 + ||B||_F^2) 
    \text{ subject to } X = AB^T 
\end{equation}
However, (\ref{eq:nnform1}) is not a convex problem since we have the multiplication of decision variables in the term $AB^T$. Equation (\ref{eq:nnform_sdp}) on the other hand is a convex formulation for the nuclear norm as a semi-definite program.
\begin{equation}
\begin{aligned}
\label{eq:nnform_sdp}
\|X\|_* \;=\; &\min_{A,\,B}\;\frac{1}{2}\Bigl(\operatorname{tr}(A) + \operatorname{tr}(B)\Bigr) \\[6pt]
&\text{subject to}\quad
\begin{bmatrix}
A & X \\
X^T & B
\end{bmatrix}
\succeq 0,\qquad
A \succeq 0,\; B \succeq 0.
\end{aligned}
\end{equation}
Let $X =U\Sigma V^T$ be the SVD of $X$. Then, the solution to the above SDP in (\ref{eq:nnform_sdp}) is given by 
$A = U \Sigma U^T$ and $B = V \Sigma V^T$.
It is a known result that the solution involves the left and right singular vectors/subspaces $U$ and $V$ of $X$. Thus, a natural question arises --- assuming the subspaces are known a priori (from observations of an adjacent day), can we inject this prior information via the parameters A and B when $X$ itself is a decision variable and still end up with a convex program? The answer turns out to be positive in general, with some constraints. We formulate such an optimization problem and its variations where $X$ is the decision variable to be imputed, given the partially observed matrix $Y$ and the mask $\Omega$. $A$ and $B$ are set to $U \Sigma U^T$ and $V\Sigma V^T$ (or approximations to them) for injecting singular subspaces priors contained in $U$ and $V$. 
\begin{equation}
    \begin{aligned}
        \min_{X} \;\; 1 \hspace{5.95cm}\\\;\\
        \text{subject to}\quad
        \begin{bmatrix}
        A & X \\
        X^T & B
        \end{bmatrix}
        \succeq 0,\qquad
        A \succeq 0,\; B \succeq 0\\
        X_{ij} = Y_{ij} \text{ if } \Omega_{ij} = 1
    \end{aligned}
    \label{eq:hresi}
\end{equation}
We can relax the constraint on $X$ to approximate the observations instead of a hard match by reformulating it as  
\begin{equation}
    \begin{aligned}
        \min_{X} || \Omega \circ (X - Y)||_F \hspace{3.8cm} \\\;\\
        \text{subject to}\quad
        \begin{bmatrix}
        A & X \\
        X^T & B
        \end{bmatrix}
        \succeq 0,\qquad
        A \succeq 0,\; B \succeq 0
    \end{aligned}
    \label{eq:sresi}
\end{equation}
In (\ref{eq:hresi}), the entries in partially observed $Y$ must exactly match the corresponding entries in our decision variable $X$ where as we relax this assumption in  (\ref{eq:sresi}) with a softer constraint. However, $A = U\Sigma U^T$ and $B = V \Sigma V^T$ are fixed which impose hard priors on the singular subspaces. (\ref{eq:srnesi-delta})  is an attempt to soften these priors in the case the subspaces being injected are approximate priors; this happens to be the case when the subspace invariance assumption holds loosely. 
\begin{equation}
    \begin{aligned}
        \min_{X,A,B} || \Omega \circ (X - Y)||_F \hspace{3.6cm} \\\;\\
        \text{subject to}\quad
        \begin{bmatrix}
        A & X \\
        X^T & B
        \end{bmatrix}
        \succeq 0,\qquad
        A \succeq 0,\; B \succeq 0\\
        ||A - U\Sigma U^T||_F \leq \delta_1 \text{ for suitable } \delta_1 \\
        ||B - V\Sigma V^T||_F \leq \delta_2 \text{ for suitable } \delta_2 \\
    \end{aligned}
    \label{eq:srnesi-delta}
\end{equation}
Alternatively, soft prior injection can also be achieved by regularizing the objective (with appropriate $\alpha$ and $\beta$) as in (\ref{eq:srnesi-regularized})
\begin{equation}
    \begin{aligned}
        \min_{X,A,B} || \Omega \circ (X - Y)||_F \; + 
        \alpha ||A - U\Sigma U^T||_F \; + \\ \beta ||B - V\Sigma V^T||_F\\\;\\
        \text{subject to}\quad
        \begin{bmatrix}
        A & X \\
        X^T & B
        \end{bmatrix}
        \succeq 0,\qquad
        A \succeq 0,\; B \succeq 0\\
    \end{aligned}
    \label{eq:srnesi-regularized}
\end{equation}
In (\ref{eq:srnesi-delta}) and (\ref{eq:srnesi-regularized}), $A,B$ are allowed to deviate in both singular values and subspaces (being loosely informed by $U$ and $V$). However, allowing flexibility exclusively in the singular values that weigh the 1-D subspaces (the columns of $U$ and $V$ are 1-D subspaces) results in the formulation (\ref{eq:srwsi}) with $d$ as the decision variable where $D := diag(d)$.
\begin{equation}
    \begin{aligned}
        \min_{X,d} || \Omega \circ (X - Y)||_F \hspace{1cm} \; \\
        \text{subject to} \quad
        \begin{bmatrix}
        A & X \\\\
        X^T & B 
        \end{bmatrix} \succeq 0,\;\;\\
        UDU^T = A \succeq 0, \quad 
        VDV^T = B \succeq 0 \\\
        \;\; d \geq 0 \qquad \qquad\\
    \end{aligned}
    \label{eq:srwsi}
\end{equation}
With respect to the notations in Table \ref{tab:notations}, $U$ and $V$, these formulations refer to $U[t_2]$ and $V[t_2]$. $\Sigma$ is $\Sigma[t_2]$. Similarly $Y$ refers to $D[t_1]$ with the mask $\Omega$ set to $\Omega[t_1]$.

\begin{table}[htbp]
\caption{Variations of SATORIS-N}
\begin{center}
\hspace{-1em}
\begin{tabular}{|c|c|c|}
\hline
\textbf{Equation} & \textbf{Method name}\\
\hline

(\ref{eq:hresi}) & Hard Reconstruction Exact Subspace Injection \texttt{(HRESI)} \\
\hline
(\ref{eq:sresi}) & Soft Reconstruction Exact Subspace Injection  \texttt{(SRESI)}\\
\hline
(\ref{eq:srnesi-delta}),(\ref{eq:srnesi-regularized}) & Soft Rec. Regularized Subspace Injection \texttt{(SRRSI)}\\
\hline
(\ref{eq:srwsi}) & Soft Reconstruction Weighted Subspace Injection \texttt{(SRWSI)} \\
\hline
Algo. \ref{alg:srisi} & Soft Rec.  Implicit Subspace Injection \texttt{(SRISI/nnmin-h)} \\
\hline
\end{tabular}
\label{tab:method_names}
\end{center}
\vspace{-1em}
\end{table}

The algorithm \ref{alg:xrysi} provides the general pseudo-code for the SATORIS-N variations.   The convex programs were implemented using the CVXPY python package\cite{diamond2016cvxpy} and the code is planned to be released as reproducible code on github, which provides an intuitive command line interface to easily reproduce our results and also apply seamlessly to other spatiotemporal datasets.

\begin{algorithm}[htbp]
  \caption{SATORIS-N: $xRySI$  $\\ x \in [H,S]$, $\;\;y \in [E, NE]$}
  \label{alg:xrysi}
  \begin{algorithmic}[1]
    \Require Incomplete matrix $D[t_1] \in \mathbb{R}^{m \times n}$, 
             binary mask $\Omega[t_1] \in \{0,1\}^{m \times n}$, 
             neighbor matrix $D[t_2] \in \mathbb{R}^{m \times n}$, 
             target rank $k$, maximum iterations $maxiter$.
    \Ensure Imputed target $\hat{D}[t_1]$.

    \State Compute truncated SVD of neighbor: $D[t_2] \approx U \Sigma V^\top$, with rank $k$.
    \State Form subspace Gram matrices: $A \gets U \, \Sigma \, U^\top$, \quad $B \gets V \, \Sigma \, V^\top$.
    \State Define decision variable $X \in \mathbb{R}^{m \times n}$; optionally $\vec d$ for (\ref{eq:srwsi})
    \State Construct block matrix:
      \[
        \text{BLOCKMAT} = \begin{bmatrix} A & X \\ X^\top & B \end{bmatrix}.
      \]
    \State Impose general constraints:
      \begin{itemize}
        \item $\text{BLOCKMAT} \succeq 0$ \quad (positive semidefinite),
        \item $A \succeq 0, B \succeq 0$
      \end{itemize}
    \State Impose specific constraints for (\ref{eq:hresi}),(\ref{eq:sresi}),(\ref{eq:srnesi-delta}),(\ref{eq:srnesi-regularized}),(\ref{eq:srwsi})
    \State Solve convex program objectives in (\ref{eq:hresi}),(\ref{eq:sresi}),(\ref{eq:srnesi-delta}),(\ref{eq:srnesi-regularized}),(\ref{eq:srwsi})
  
    \State Let $\hat{D}[t_1]_{ij} \gets X^\star_{ij}$(the optimizer) if $\Omega[t_1]_{ij} = 1$
    \State \Return $\hat{D}[t_1]$.
  \end{algorithmic}
\end{algorithm}

\begin{algorithm}[htbp]
  \caption{SATORIS-N: $SRISI/NNmin-h$}
  \label{alg:srisi}
  \begin{algorithmic}[1]
    \Require Incomplete matrix $D[t_1] \in \mathbb{R}^{m \times n}$, 
             binary mask $\Omega[t_1] \in \{0,1\}^{m \times n}$, 
             neighbor matrix $D[t_2] \in \mathbb{R}^{m \times n}$, 
             target rank $k$, maximum iterations $maxiter$.
    \Ensure Imputed target $\hat{D}[t_1]$.

    \State Stack observations: $Y := \big[ \; D[t_1] \; | \; D[t_2] \; \big]$
    \State Stack masks: $\Omega := \big[ \;  \Omega[t_1] \; | \; 11^T \; \big] $
    
    \State Solve convex program to get optimal $X^*$: 
    $$\min_{X} || \Omega \circ (X - Y)||_F + ||X||_*$$

    \State Let $\hat{D}[t_1]_{ij} \gets X^\star_{ij}$ if $\Omega[t_1]_{ij} = 1$
    \State \Return $\hat{D}[t_1]$.
  \end{algorithmic}
  
\end{algorithm}

\section{Experiments}
\label{sec:experiments}

\subsection{Preliminary Experiments on Images}
\begin{figure}[h]
\centering
\centerline{\includegraphics[trim=2.4cm 2cm 2cm 2cm,clip,width=25em]{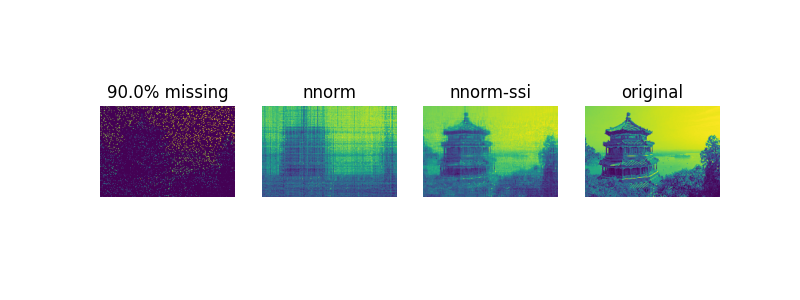}}
\centerline{\includegraphics[trim=2.4cm 2.5cm 2cm 2.5cm,clip,width=25em]{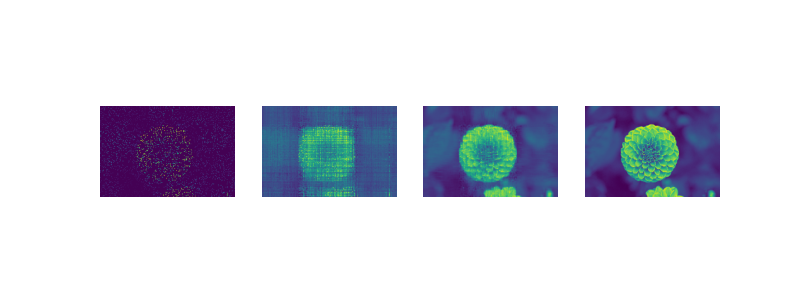}}
\vspace{-1em}
\caption{Image imputation using vanilla nuclear norm minimization and subspace informed nuclear norm minimization. \emph{nnorm} is vanilla nuclear norm minimization and \emph{nnmorm-ssi} is subspace informed nuclear norm minimization using hard reconstruction (\texttt{HRESI}). Images are from sklearn load\_sample\_images().}
\label{fig:prelim_images}
\vspace{-1em}
\end{figure}

Before applying our method to traffic data, we conducted a preliminary
image-completion experiment to assess the practical value of subspace-informed
nuclear-norm minimization. As motivated in
Section~\ref{sec:problemformulation}, the SDP formulation
(\ref{eq:nnform_sdp}) encourages reconstructions aligned with known
left/right singular subspaces. To test this effect in a simple setting, we
performed matrix completion on images using either (i) the vanilla nuclear
norm or (ii) our explicit subspace-informed variant, where the subspaces are
the left and right singular vectors obtained from the SVD of the images.
Figure~\ref{fig:prelim_images} shows that incorporating subspace information
improves reconstruction quality compared to uninformed formulation,
providing an early empirical demonstration of its usefulness.

\subsection{Experiment Setup for Traffic Data}
\begin{figure}[h]
    \centering
    \includegraphics[width=\columnwidth]{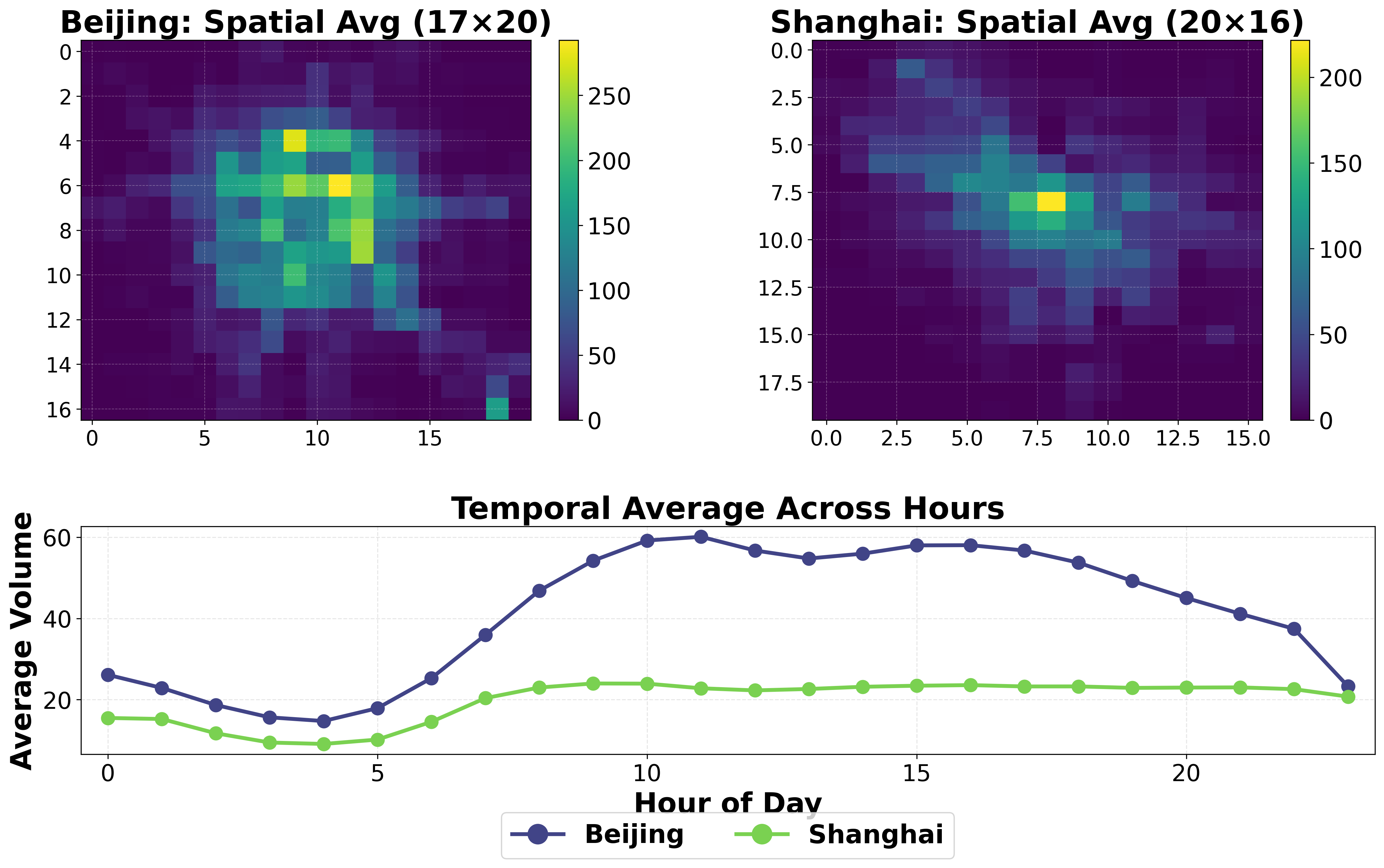}
    \vspace{-2em}
    \caption{Spatial and temporal averages over a single 24-hour day in the Beijing and Shanghai datasets.}
    \label{fig:dataexplore-spatiotemporal}
    \vspace{-1em}
\end{figure}

We evaluated our methods on two real-world taxi-GPS datasets from Beijing (BEI)
and Shanghai (SHANG), represented as daily traffic-density matrices of size
$340\times24$ and $320\times24$, respectively (rows: sensor locations;
columns: hourly time slots). Each entry is an hourly aggregated vehicle
density; the resulting matrices exhibit strong spatiotemporal regularities
and are natural candidates for low-rank recovery. Data are used in their raw
(unnormalized) form. To simulate missingness, we randomly mask entries at
five levels (10\%, 25\%, 50\%, 75\%, 90\%). For each level, experiments are
run over seven days, and we report the mean and standard deviation across
days. Figure~\ref{fig:dataexplore-spatiotemporal} shows spatial (top) and
temporal (bottom) averages over 24 hours for both cities.

We compare ten imputation methods: statistical
(\texttt{mean}, \texttt{KNN}, \texttt{KNNW}, \texttt{MICE}~\cite{MICE},
\texttt{MissForest}~\cite{MissForest}), low-rank (\texttt{SoftImpute},
\texttt{IterativeSVD}, \texttt{NNmin}~\cite{fancyimpute}), and deep-learning
(\texttt{GAIN}~\cite{GAIN}, \texttt{MIRACLE}~\cite{MIRACLE,Jarrett2022HyperImpute}).
For each method, we also evaluated two implicit subspace–informed variants by
concatenating an adjacent day either horizontally (\texttt{-h}, promoting
shared column-space/spatial modes) or vertically (\texttt{-v}, promoting
shared row-space/temporal modes), resulting in 30 configurations per missingness
level. We emphasize that \texttt{NNmin-h} is our proposed implicitly informed
nuclear-norm model, \emph{Soft Reconstruction Implicit Subspace Injection}
(\texttt{SRISI}).

\subsection{Temporal Stability of Singular Subspaces}
\label{sec:subspace_stability}

To justify subspace-informed imputation, we first quantify how stable the
singular-vector subspaces are across adjacent days in the Beijing and
Shanghai datasets. We use a \emph{subspace overlap} metric defined as the
mean of the top-$k$ singular values $\sigma_i$ of $U_1^\top U_2$, where
$U_1, U_2 \in \mathbb{R}^{n\times k}$ are the left or right singular-vector
matrices for consecutive days:
\begin{equation}
\label{eq:subspace_overlap_metric}
    \bar\sigma = \frac{1}{k}\sum_{i=1}^k \sigma_i, \qquad
    \text{std}(\sigma) = \sqrt{\frac{1}{k}\sum_{i=1}^k(\sigma_i - \bar\sigma)^2}.
\end{equation}
This metric lies in $[0,1]$, with $1$ indicating identical subspaces and $0$
indicating orthogonality; higher values correspond to stronger cross-day
alignment and greater potential for effective subspace transfer.

\begin{figure}[h]
\centering
\includegraphics[width=25em]{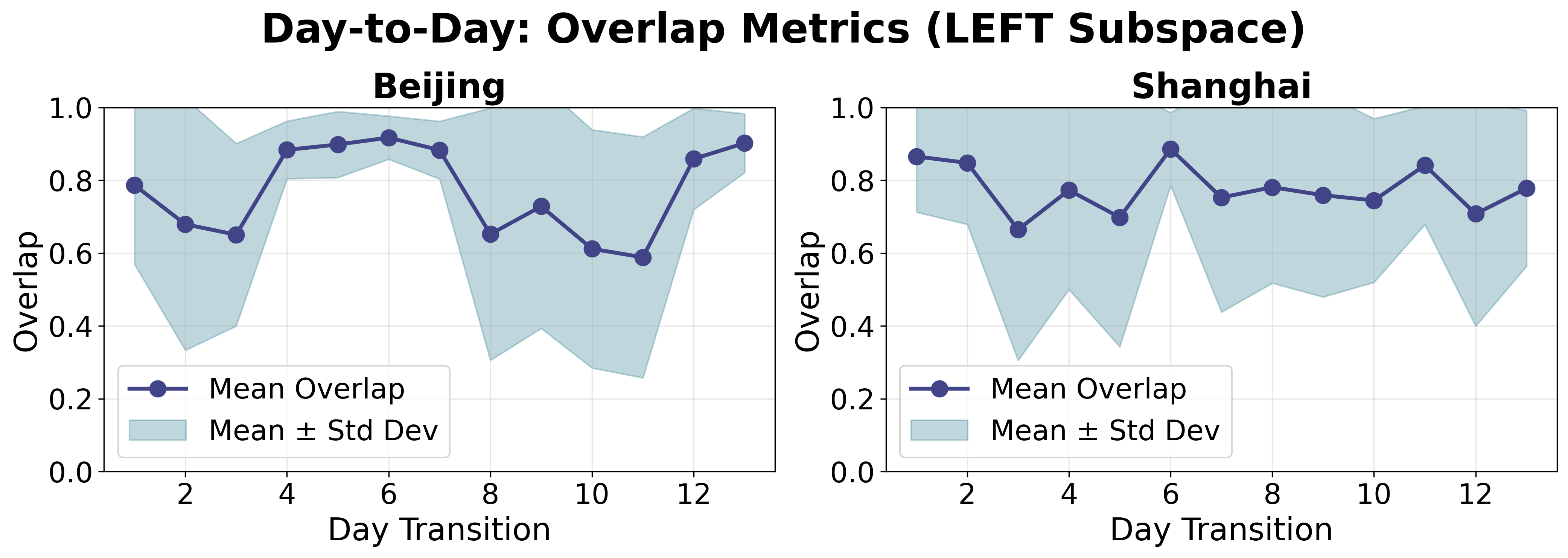}
\includegraphics[width=25em]{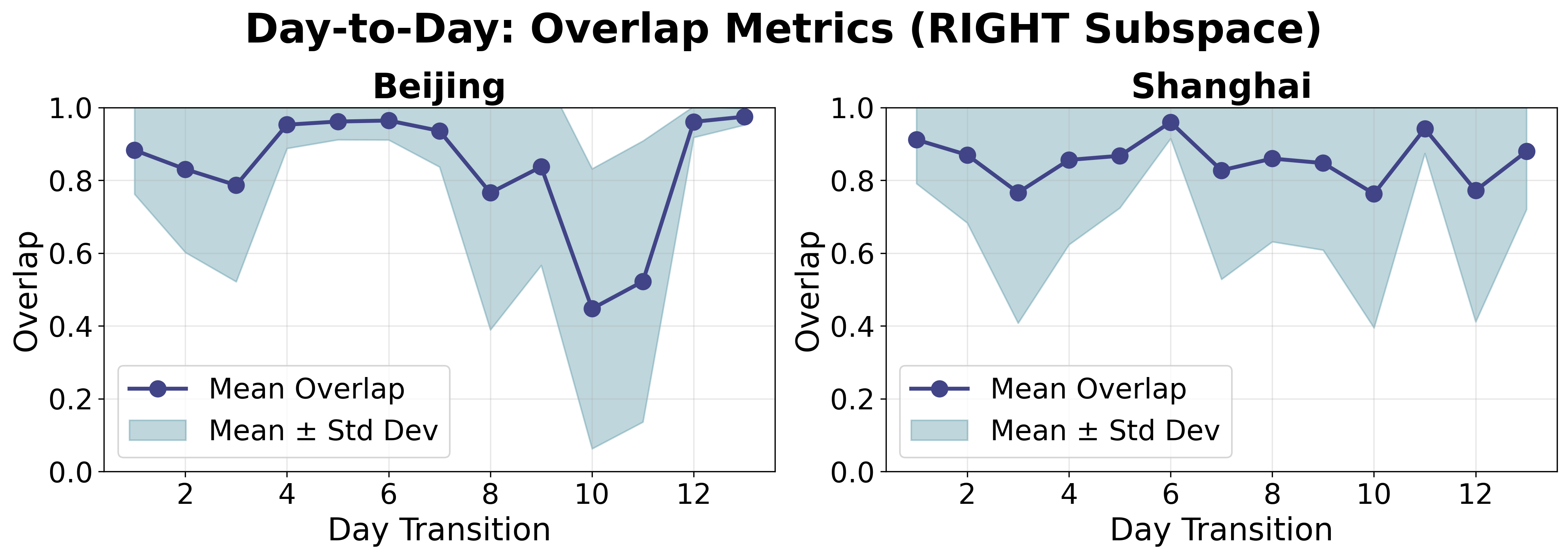}
\vspace{-1.5em}
\caption{Stability of left and right singular subspaces for Beijing and
Shanghai across adjacent days using the subspace-overlap metric in
(\ref{eq:subspace_overlap_metric}); $0$ = orthogonal, $1$ = perfectly aligned.}
\label{fig:subspace_stability}
\vspace{-2em}
\end{figure}

Figure~\ref{fig:subspace_stability} shows the mean and one-standard-deviation
bands of subspace overlap for left and right singular subspaces across
adjacent days. Both datasets exhibit substantial overlap (values much closer
to $1$ than $0$), with Shanghai showing higher and more consistent alignment
than Beijing, consistent with the stronger imputation performance observed
later.

\subsection{Implicit Subspace Alignment Findings}

\begin{figure}[h]
    \centering
     
    \includegraphics[width=\columnwidth,trim={0 3.9cm 0 0cm},clip]{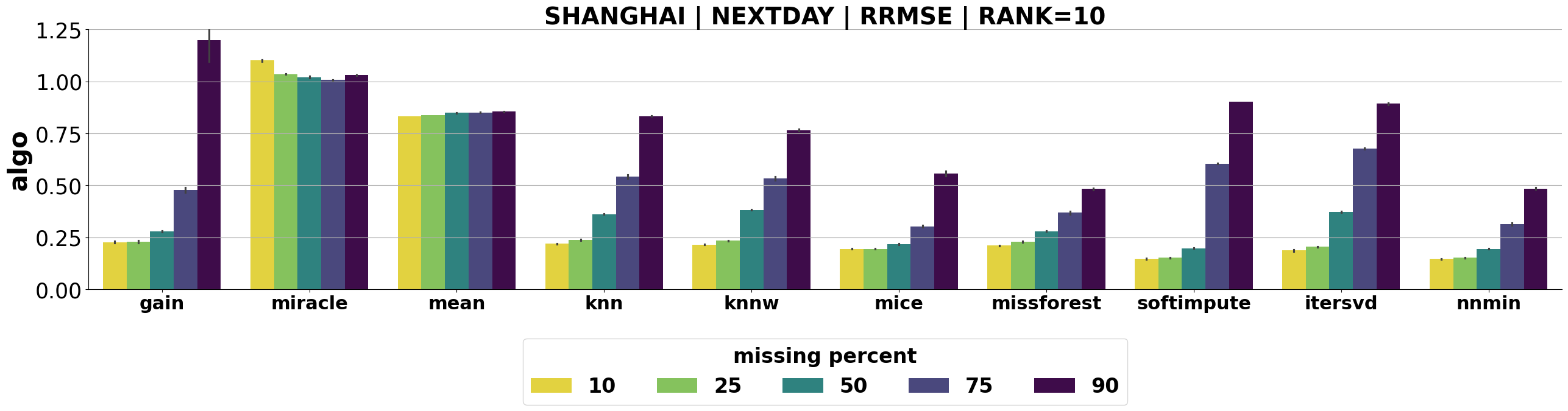}
    \includegraphics[width=\columnwidth,trim={0 3.9cm 0 1.3cm},clip]{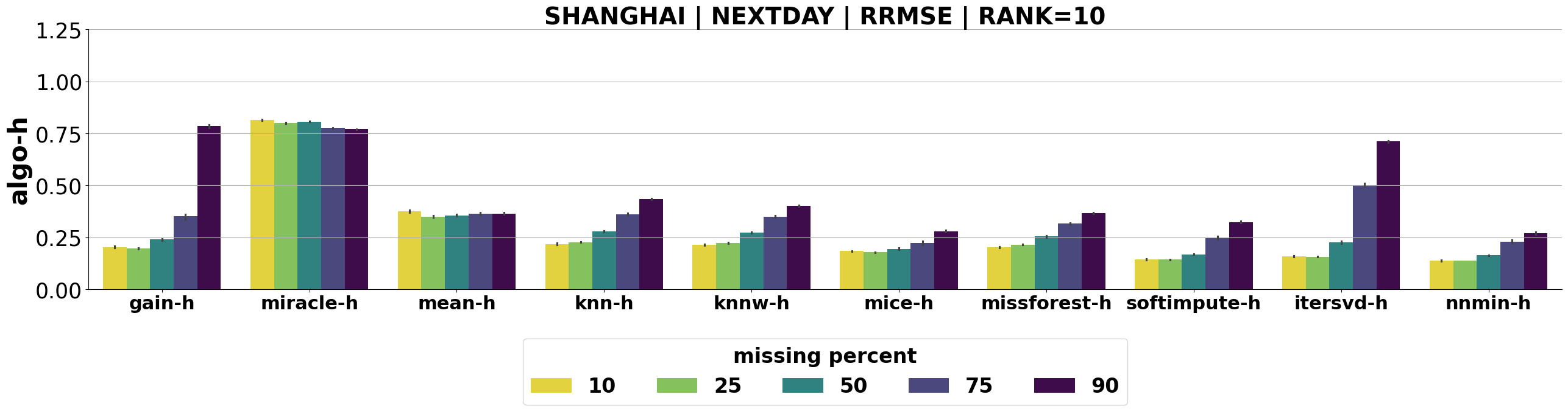}
    \includegraphics[width=\columnwidth,trim={0 3.9cm 0 1.3cm},clip]{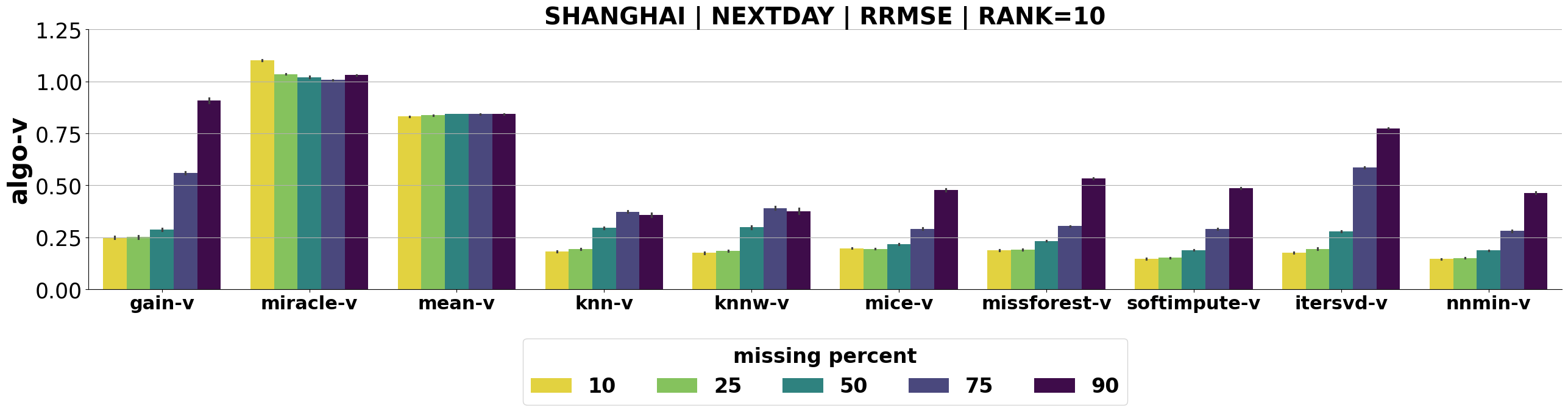}

    \vspace{1em}

     \includegraphics[width=\columnwidth,trim={0 3.9cm 0 0cm},clip]{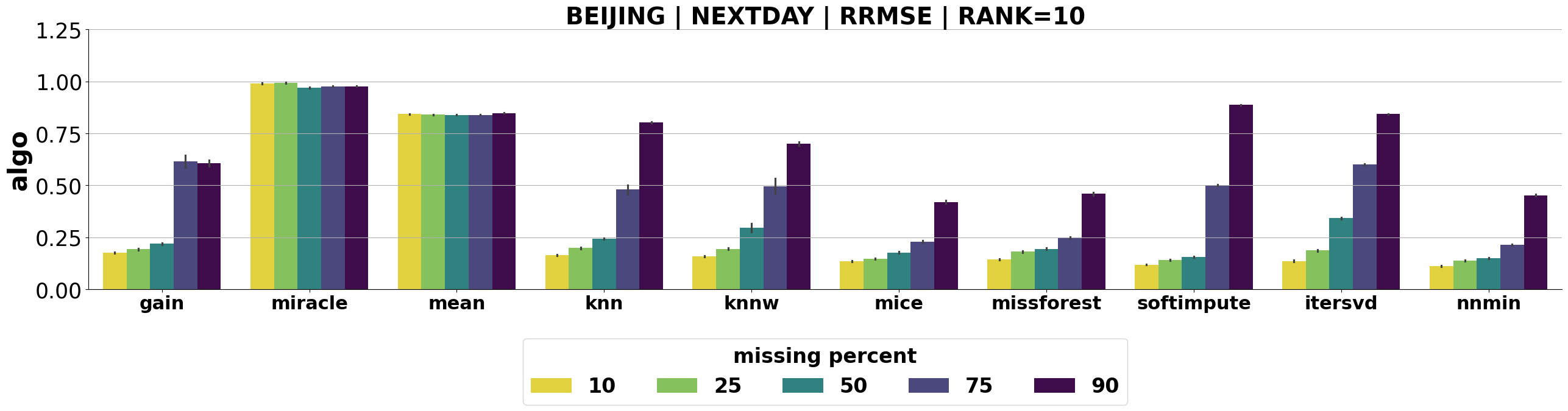}
    \includegraphics[width=\columnwidth,trim={0 3.9cm 0 1.3cm},clip]{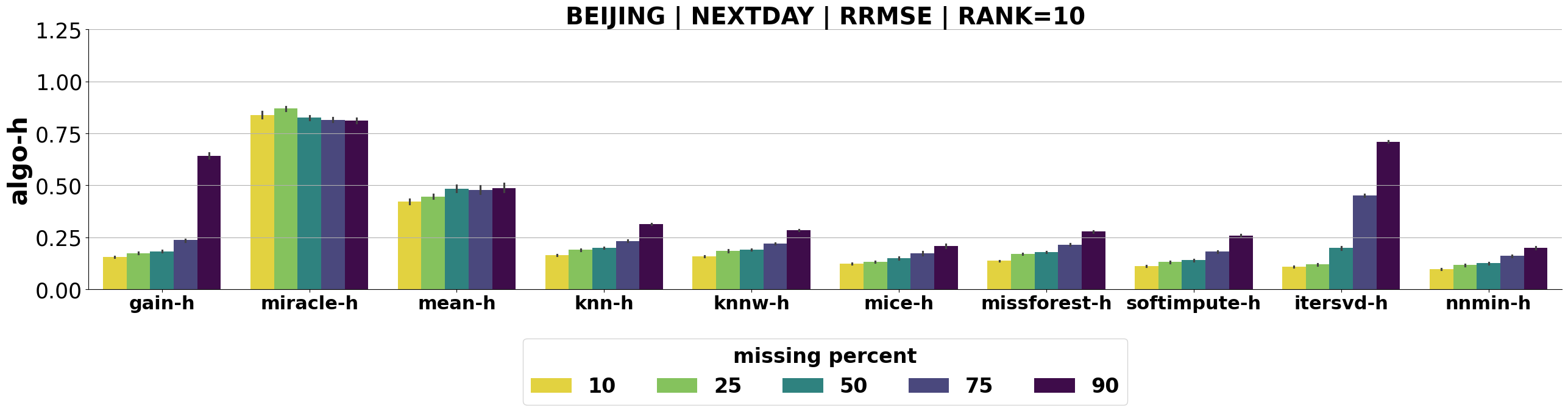}
    \includegraphics[width=\columnwidth,trim={0 0 0 1.3cm},clip]{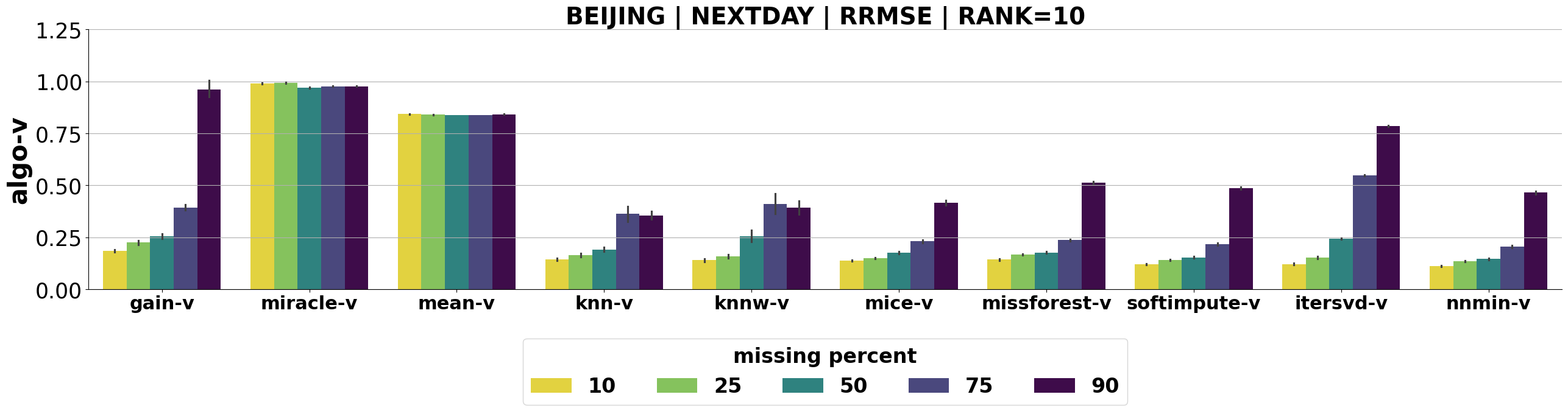}

    \vspace{-1em}
    \caption{Comparison of Relative Root Means Square Error (RRMSE) for 10 imputation algorithms and their implicit subspace-informed variants across varying missing percentages (mp) for Shanghai dataset. The top row is the baseline, the second the \texttt{-h} (proposed horizontal stacking) and the last \texttt{-v} (proposed vertical stacking) variant. Implicit priors yields performance gains, especially under moderate to high occlusion. Gains are more pronounced for vertical  stacking (-h) since that allows shared spatial subspaces v.s temporal in horizontal stacking (-v). Highlighted \texttt{NNmin-h/SRSI} is comparable and relates to our other explicitly informed subspace-aware nuclear norm based imputation methods}
    
    \label{fig:compare_algo_variants_shang_bei}
\end{figure}

We assess implicit subspace alignment by comparing each baseline algorithm
with its horizontally (\texttt{-h}) and vertically (\texttt{-v}) stacked
variants (Fig. ~\ref{fig:compare_algo_variants_shang_bei}). Three main trends emerge. First,
stacking direction matters: horizontal stacking (\texttt{-h}) typically
yields the largest gains, which is consistent with promoting alignment in
the higher-dimensional spatial subspace (320 for Shanghai, 340 for Beijing)
rather than the 24-dimensional temporal subspace. Second, all methods—
especially low-rank models such as \texttt{SoftImpute}, \texttt{IterativeSVD},
and \texttt{NNmin}—show clear improvements under high missingness (75–90\%)
when augmented with implicit alignment, indicating that side-channel
information from neighboring days enhances subspace consistency and
robustness. Third, deep-learning models (\texttt{MIRACLE}, \texttt{GAIN})
improve only at low occlusion and degrade sharply as missingness increases,
reflecting their higher sample complexity. Classical methods such as
\texttt{NNmin}, \texttt{MICE}, \texttt{SoftImpute}, and \texttt{MissForest}
perform reliably with and without stacking. Overall, these results
support the hypothesis that even simple forms of subspace coupling via
concatenation can substantially improve imputation at high missing rates
without changing the underlying optimization algorithms.

\subsection{Explicit Subspace-Aware Methods}
\begin{figure}[h]
    \centering
    \includegraphics[width=\columnwidth]{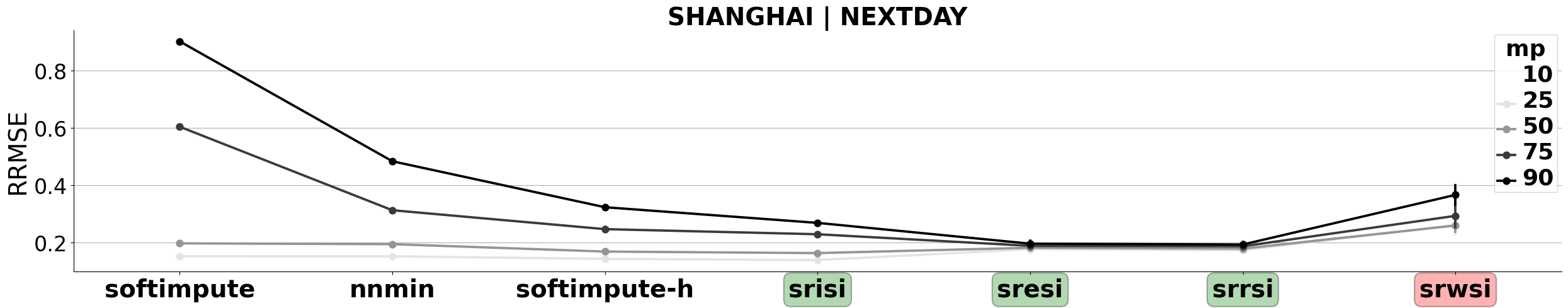}
    \includegraphics[width=\columnwidth]{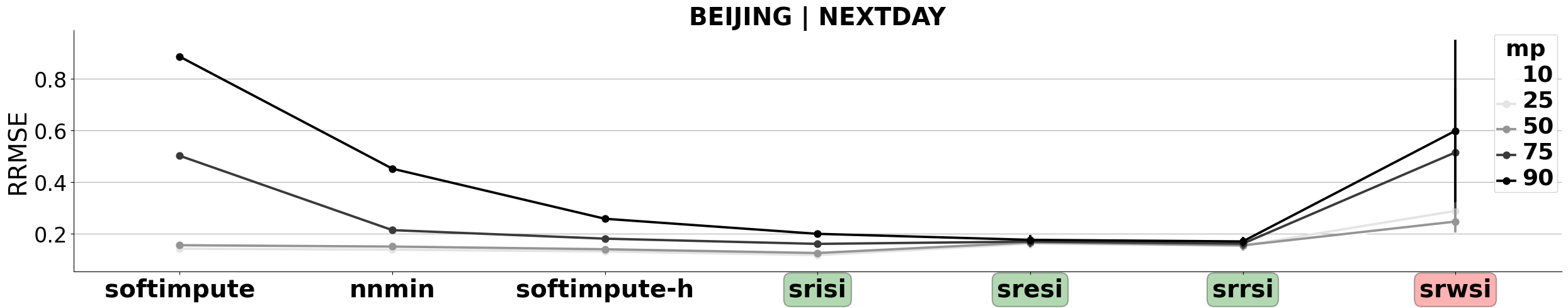}

    \vspace{-1em}
    \caption{RRMSE comparison within proposed nuclear norm based algorithms: Explicit subspace-informed methods \texttt{SRESI},
    \texttt{SRESI}, \texttt{SRRSI}, and \texttt{SRWSI} with other nuclear norm methods like \texttt{NNmin}, \texttt{SoftImpute/Softimpute-h}  on Shanghai (top)
    and Beijing (bottom). All our methods, except \texttt{SRWSI}, improve imputation performance at high occlusion levels (above 75\%).}
\vspace{-1.5em}
    \label{fig:compare_explicit_ssi}
\end{figure}

We next evaluate our explicit subspace-aware imputation algorithms—
\texttt{SRESI}, \texttt{SRRSI}, and \texttt{SRWSI}—which inject estimated
singular-subspace priors into a convex semidefinite program. As shown in
Figure~\ref{fig:compare_explicit_ssi}, \texttt{SRESI} and \texttt{SRRSI}
achieve very competitive performance in all settings, while \texttt{SRWSI}
consistently underperforms. Consequently, we focus on \texttt{SRESI} and
\texttt{SRRSI} in the subsequent results and comparisons. In preliminary experiments, we also found that using the rank k = 10
for the extraction of the subspace from the nearly complete neighbor day
works well in both datasets. We selected k through grid search
on \{1, 3, 5, 10, 15, 20, 23\} based on validation performance. For SRRSI,
we set $\beta = \alpha = 1$ after searching over {0.001, 0.01, 0.1, 1, 5, 10}.
We fix these values in all remaining explicit subspace-aware experiments.

\section{Results}
\label{sec:results}

\begin{figure}[h]
    \centering
    \includegraphics[width=\columnwidth,trim={0 3.7cm 0 0cm},clip]{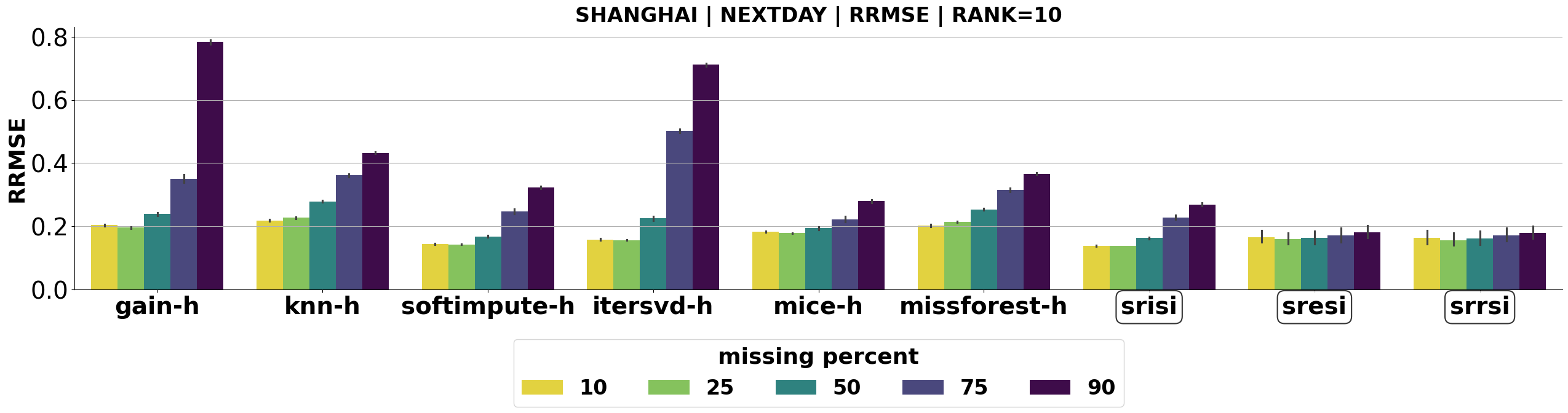}
    \includegraphics[width=\columnwidth,trim={0 3.7cm 0 0cm},clip]{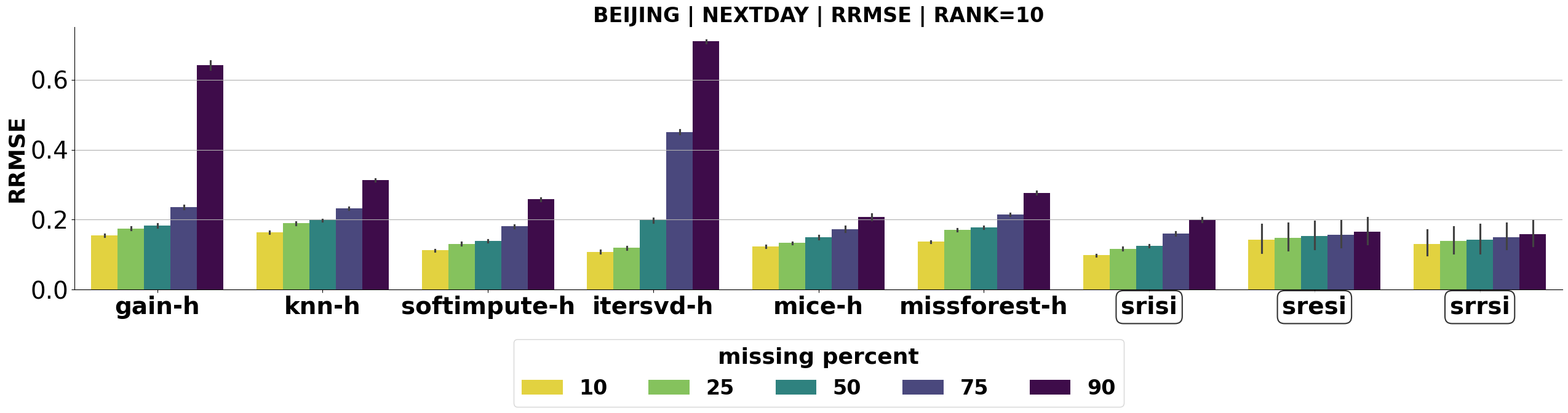}
    
    \includegraphics[width=\columnwidth,trim={0 3.7cm 0 0cm},clip]{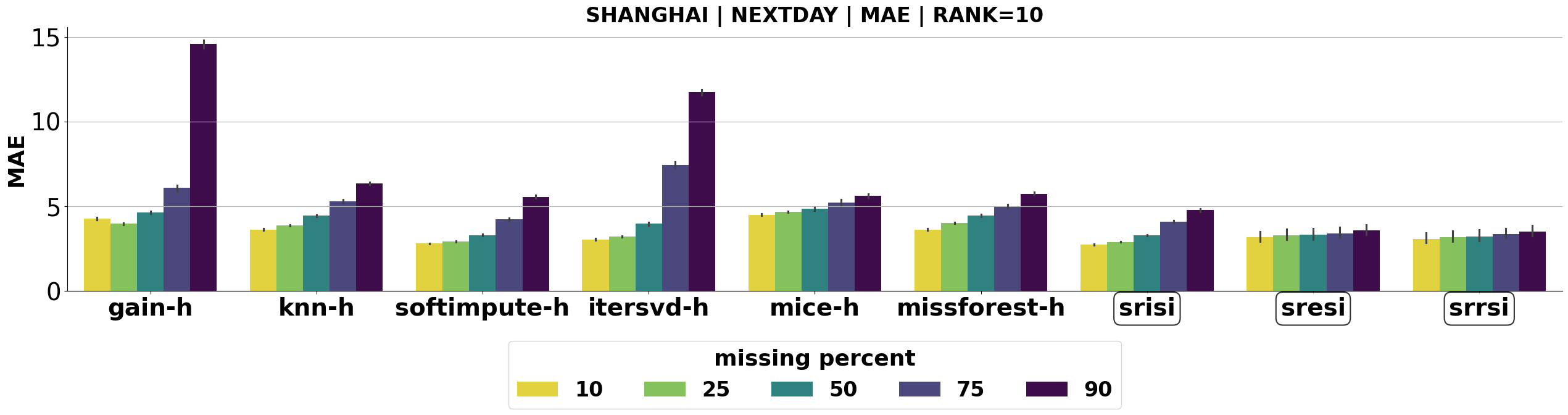}
    \includegraphics[width=\columnwidth]{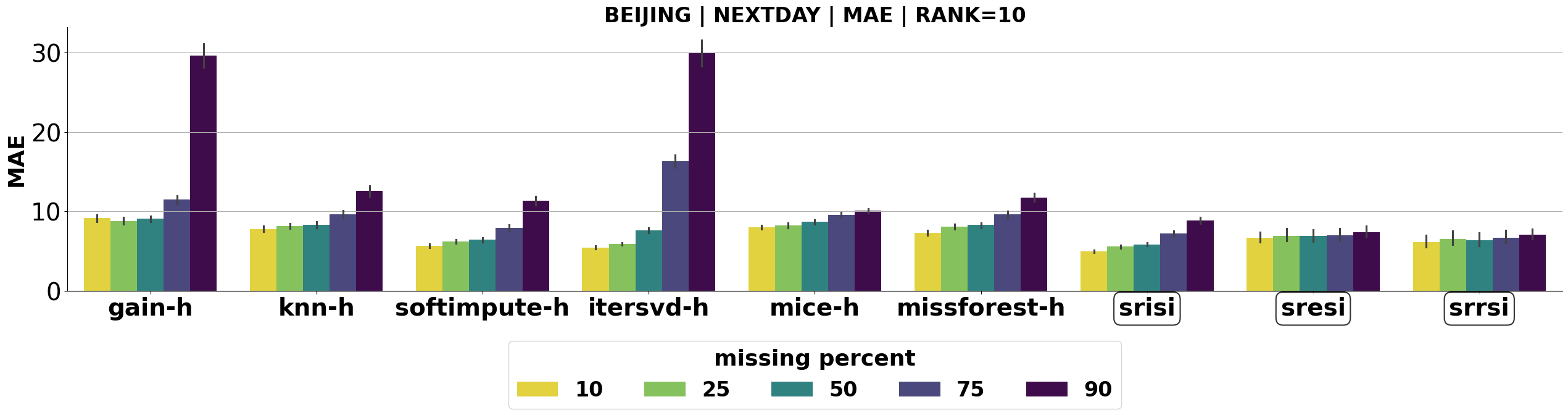}
    
\vspace{-1em}
\caption{RRMSE and MAE for proposed methods (rightmost three) and
    baselines across missingness levels over seven days. Subspace rank 
    $k = 10$ is selected via grid search over $\{1, 3, 5, 10, 15, 20, 23\}$ using 
    the next-day observation matrix. For \texttt{SRRSI}, $\alpha = \beta = 1$ 
    after searching over $\{0.001, 0.01, 0.1, 1, 5, 10\}$. Explicit methods
    (\texttt{SRESI}, \texttt{SRRSI}) perform best at high occlusion; implicit
    \texttt{SRISI/NNmin-h} performs best at low occlusion.}
    \label{fig:result_comprehensive}
    \vspace{-2em}
\end{figure}

We compare the strongest implicit
subspace-informed baselines (\texttt{KNN-h}, \texttt{SoftImpute-h},
\texttt{IterativeSVD-h}, \texttt{MICE-h}, \texttt{MissForest-h},
\texttt{GAIN-h}, and our \texttt{SRISI/NNmin-h}) to the explicit methods
\texttt{SRESI} and \texttt{SRRSI}. RRMSE and MAE results for both datasets are
shown in Figure~\ref{fig:result_comprehensive}.
\textbf{Explicit methods:} \texttt{SRESI} and \texttt{SRRSI} perform
essentially identically and achieve the lowest errors across all missingness
levels. Their advantage is largest at extreme sparsity (75--90\%), reflecting
more effective use of explicit subspace priors. At low occlusion, however,
these priors can introduce bias because the target-day subspaces can be
recovered directly from dense observations.
\textbf{Implicit method:} Our proposed \texttt{SRISI/NNmin-h} is competitive throughout
and performs best at low missingness, consistent with nuclear-norm
minimization adapting naturally to available data. \texttt{MICE-h} also
remains strong across all occlusion levels.
\textbf{Other baselines:} Deep-learning models (\texttt{GAIN-h},
\texttt{MIRACLE}) improve when observations are dense but degrade sharply
beyond 75\% missingness, confirming their sensitivity to sparse inputs.
Classical and low-rank baselines (\texttt{KNN-h}, \texttt{SoftImpute-h})
perform well at 10--25\% missingness, but deteriorate rapidly after 50\%,
especially in Shanghai. \texttt{SoftImpute} is competitive at low and
moderate sparsity, but still lags behind explicit methods at high occlusion,
despite its close relation to nuclear-norm minimization (iterative
soft-thresholding vs.\ direct convex optimization). Tree-based
\texttt{MissForest-h} performs moderately up to ~75\% missingness before
dropping off.
\textbf{Dataset effects:} Trends are consistent across Beijing and Shanghai,
though gains are larger in Shanghai, which exhibits higher adjacent-day
subspace stability (Fig.~\ref{fig:subspace_stability}). This supports the
utility of transferable subspace priors in real traffic matrices.
\textbf{Practical takeaway:}  
\emph{Use implicit subspace priors (\texttt{SRISI/NNmin-h}) when missingness
is low, and explicit priors (\texttt{SRESI} or \texttt{SRRSI}) when
missingness is high. This mixed strategy outperforms all baselines.}
The explicit methods \texttt{SRESI} and \texttt{SRRSI} rely on semidefinite programming (SDP) solvers, which incur higher computational cost than iterative approaches, though they remain practical for offline or periodic traffic reconstruction.

\section{Conclusion and Future Work}
\label{sec:conclusion}

This paper introduced \textbf{SATORIS-N}, a subspace-aware traffic imputation
framework that leverages both implicit and explicit singular-subspace priors
to address missing data in spatiotemporal traffic matrices. Exploiting the
temporal stability of singular subspaces across adjacent days, we showed that
reusing prior subspace information—either implicitly via matrix stacking or
explicitly via a convex nuclear-norm SDP—yields substantial gains over
conventional low-rank and learning-based baselines. The explicit SDP-based
methods (\texttt{SRESI}, \texttt{SRRSI}) align reconstructions with known
subspaces and are particularly robust under extreme occlusion (up to 90\%
missing entries), while the lightweight implicit approach 
offers a practical drop-in meta-algorithm can enhance standard matrix-imputation
methods with zero code modification. Across two real-world datasets, our informed
methods consistently outperform both classical and deep learning baselines,
highlighting the value of convex, subspace-informed modeling for traffic
completion. The proposed framework is particularly relevant for intelligent vehicle 
systems. In V2X networks with a sparse penetration of CV (10--30\%), SATORIS-N 
can reconstruct complete traffic density fields from incomplete vehicle 
reports, allowing cooperative perception and predictive routing for 
autonomous vehicles. Future work will evaluate real-time performance on 
embedded platforms and integration with V2X communication protocols.

Several directions arise from this work. \emph{Theoretical generalizations}
include incorporating additional structure (graph smoothness, periodicity,
sparsity) into the convex SATORIS-N framework, drawing on probabilistic
subspace models such as BayeSMG~\cite{bayeSMG2021} and robust PCA-style
decompositions~\cite{PriorMaxCorr2018}, as well as establishing sample
complexity and recovery guaranties. \emph{Scalability and real-time
deployment} motivate large-scale SDP approximations (e.g. 
SketchyCGAL~\cite{sketchyCGAL2021}, first-order randomized methods),
distributed schemes (e.g. ADMM), warm-start strategies, and adaptive
updates to singular values for latency-sensitive tasks such as congestion
detection. \emph{Dynamic subspace modeling} is another promising avenue:
online low-rank methods such as GROUSE and GRASTA~\cite{grouse2014},
sliding-window PCA, and Kalman-style filters could help SATORIS-N track
seasonal changes, policy shifts, and anomaly-induced regime changes. 

\emph{Hybrid convex–deep architectures} offer opportunities to combine
interpretability with flexibility—for example, via learned SVT
unrollings~\cite{learnedSVT2021}, GNN or autoencoder-based nonlinear priors,
or neural surrogate solvers that accelerate convex inference while
preserving structure. Extending SATORIS-N to \emph{multimodal and contextual}
settings by fusing exogenous signals (e.g. weather, events, social media)
builds on prior work showing robustness gains from external covariates
in extreme conditions~\cite{weatheraware2020}, and suggests tensor-based
extensions with auxiliary channels. Finally, \emph{societal deployment and
smart-city integration} remain important: embedding SATORIS-N into urban
analytics platforms or digital twins can support transit planning,
infrastructure management, and equity-aware sensing, complementing
data-driven smart-city efforts that have already demonstrated reductions in
congestion and emissions~\cite{bigDataSmartCity2025}.

\vspace{-1em}
\section*{Acknowledgment}
The authors thank Dr. Fan Bai of General Motors Research for providing the traffic data sets utilized in this investigation.

\bibliographystyle{IEEEtran}           
\bibliography{refs} 

\end{document}